\documentclass[paper]{ieice}
\usepackage[pdftex]{graphicx,xcolor}
\usepackage[fleqn]{amsmath}
\usepackage{cite}
\usepackage{hyperref}
\usepackage{times} 
\usepackage{indentfirst} 
\usepackage{tabularx}
\newcolumntype{Y}{>{\centering\arraybackslash}X}
\usepackage{amssymb}
\usepackage{booktabs} 
\usepackage{multirow} 
\usepackage[table]{xcolor}
\usepackage{colortbl}
\definecolor{sampleOne}{RGB}{235,245,255}  
\definecolor{sampleThree}{RGB}{235,255,235} 
\definecolor{sampleFive}{RGB}{255,245,235}  
\field{}

\title{DMP-3DAD: Cross-Category 3D Anomaly Detection via Realistic Depth Map Projection with Few Normal Samples}
\authorlist{%
\authorentry{Zi Wang}{n}{niigata}\MembershipNumber{}
\authorentry{Katsuya Hotta}{n}{iwate}\MembershipNumber{}
\authorentry{Koichiro Kamide}{n}{toyama}\MembershipNumber{}
\authorentry{Yawen Zou}{n}{toyama}\MembershipNumber{}
\authorentry{Jianjian Qin}{n}{ningbo}\MembershipNumber{}
\authorentry{Chao Zhang}{m}{toyama}\MembershipNumber{1508946}
\authorentry{Jun Yu}{n}{niigata}\MembershipNumber{}
}
\affiliate[niigata]{Niigata University}
\affiliate[toyama]{University of Toyama}
\affiliate[iwate]{Iwate University}
\affiliate[ningbo]{NingboTech University}

\received{2024}{1}{1}
\revised{2024}{1}{1}

\begin{document}
\maketitle
\begin{summary}
Cross-category anomaly detection for 3D point clouds aims to determine whether an unseen object belongs to a target category using only a few normal examples.
Most existing methods rely on category-specific training, which limits their flexibility in few-shot scenarios.
In this paper, we propose DMP-3DAD, a training-free framework for cross-category 3D anomaly detection based on multi-view realistic depth map projection. Specifically, by converting point clouds into a fixed set of realistic depth images, our method leverages a frozen CLIP visual encoder to extract multi-view representations and performs anomaly detection via weighted feature similarity, which does not require any fine-tuning or category-dependent adaptation.
Extensive experiments on the ShapeNetPart dataset demonstrate that DMP-3DAD achieves state-of-the-art performance under few-shot setting. The results show that the proposed approach provides a simple yet effective solution for practical cross-category 3D anomaly detection.
\end{summary}


\section{Introduction}

Cross-category anomaly detection for 3D point clouds \cite{qin2023teacher, masuda2021toward} is an important but sparsely studied problem in 3D vision. In many practical settings, one may only have access to a few normal exemples of a target category, yet the system must identify whether an unseen sample belongs to the same category or a different/unknown category. Unlike category-specific anomaly detection \cite{bergmann2021mvtec,liu2023real3d} (e.g., surface defect detection on a known category), cross-category anomaly detection requires distinguishing semantic differences across object classes. Moreover, point clouds present challenges such as sparsity, irregular sampling, and viewpoint variations, which make robust anomaly scoring challenging.

Existing 3D anomaly detection methods are generally built upon training mechanisms that require training normal examples for model optimization. Reconstruction-based approaches such as variational autoencoders aim to learn a compact representation of normal data and identify deviations at test time \cite{masuda2021toward}. Similarly, teacher–student and distillation frameworks model the normal feature distribution \cite{qin2023teacher}. However, these training-based methods must be retrained or fine-tuned when the normal category changes, which limits their flexibility in few-shot scenarios. Recently, CLIP-based multi-view frameworks have been explored for zero-shot 3D anomaly detection using auxiliary supervision or prompt constructs \cite{cheng2024towards}, but such methods typically depart from the strict one-class setting because of introducing additional training components with external data.

A possible strategy to address cross-category anomaly detection is to leverage the power of large pretrained 2D visual models via multi-view representations of 3D data. Multi-view projection techniques \cite{zhu2023pointclip,zuo2024clip3d} convert point clouds into sets of 2D images, enabling the use of strong 2D backbones without designing specialized 3D networks. In particular, realistic depth map projection has been introduced to produce depth renders that better match the distribution of images seen by 2D vision models \cite{zhu2023pointclip}. When combined with a pretrained CLIP visual encoder, such multi-view representations provide rich, semantically meaningful embeddings that can support cross-category discrimination without retraining.

Inspired by this, in this paper, we propose DMP-3DAD, a training-free framework for cross-category 3D point cloud anomaly detection. Given only a few normal reference point clouds from a target category, our approach does not require any training data preparation, optimization, or category-specific adaptation. We first project point clouds into a fixed set of multi-view realistic depth maps based on an established rendering strategy \cite{zhu2023pointclip}, extract feature embeddings using a frozen CLIP visual encoder \cite{radford2021learning}, and then compute weighted view-wise similarities between the test sample and the normal references. At last, by aggregating view-specific scores, our method yields a robust anomaly score that can distinguish unseen categories under few-shot conditions. Our contributions can be concluded threefold:
\begin{itemize}
    \item We introduce a training-free solution for cross-category 3D anomaly detection on point clouds, enabling anomaly scoring without model training or category-wise adaptation when the normal class changes.  
    \item We leverage multi-view realistic depth map projection and a frozen CLIP visual encoder to obtain robust representations from sparse 3D point clouds, forming a simple yet effective pipeline that generalizes across categories.
    \item We demonstrate that the proposed approach achieves state-of-the-art performance under a few-shot setting, outperforming prior training-based methods.
\end{itemize}

\section{Related Work}
\label{sec:related_work}

\subsection{3D Point Cloud Anomaly Detection}
\label{sec:rw_3dad}

\noindent\textbf{Category-specific 3D anomaly detection.}
Category-specific 3D anomaly detection is mainly studied in industrial inspection scenarios, where anomalies are defined as local defects under a fixed object category.
Two representative benchmarks have largely developed this research direction. MVTec 3D-AD \cite{bergmann2021mvtec} extends the widely used 2D anomaly detection benchmark to 3D by providing depth-based data with fine-grained defect annotations \cite{bergmann2021mvtec}.
More recently, Real3D-AD \cite{liu2023real3d} further extend this line by introducing high-resolution point cloud data and systematic evaluations, highlighting challenges such as precise geometry modeling and subtle defect patterns \cite{liu2023real3d}.
Motivated by these benchmarks, most existing methods focus on reconstructing or modeling normal geometry and detecting deviations in feature or reconstruction spaces, often incorporating geometry-aware designs, distillation mechanisms, or multimodal cues \cite{cheng2025mc3d,wang2023multimodal,horwitz2023back,wang20253dkeyad, wang2025boosting}.
Overall, category-specific 3D anomaly detection mainly focus on defect-level abnormality and fine-grained localization on fixed categories, which is fundamentally different from the cross-category setting considered in our work. Interested readers are referred to recent survey articles for a comprehensive overview~\cite{du20253d,liu2024deep,lin2025survey}.

\noindent\textbf{Cross-category 3D anomaly detection.}
Cross-category 3D anomaly detection aims to distinguish normal samples of a target category from anomalous samples originating from unseen categories, which is commonly formulated as one-class and few-shot anomaly detection at the object level.
In this setting, only normal samples (often very few) are available for the target category during training or reference construction, and the detector must reject out-of-category samples at test time.
Representative approaches typically model the distribution of normal point clouds and identify anomalies as deviations in reconstruction or feature space.
Masuda \emph{et al.} propose a variational autoencoder to learn compact normal representations and detect anomalies via reconstruction errors \cite{masuda2021toward}.
Qin \emph{et al.} propose a teacher-student framework in which the teacher and student networks exhibit different sensitivities to normal feature patterns, and anomalies are detected by measuring the discrepancy between their feature representations~\cite{qin2023teacher}.
While effective, these methods rely on category-dependent training procedures and usually require re-training or fine-tuning when the normal category changes, which limits their flexibility in practical cross-category scenarios.

\subsection{CLIP-based Anomaly Detection}
\label{sec:rw_clip_ad}

CLIP \cite{radford2021learning} has demonstrated strong generalization ability and has recently been adopted for anomaly detection.
In 2D, several works adapt CLIP via prompt learning or anomaly-aware alignment to distinguish normal and abnormal patterns without task-specific training \cite{zhou2023anomalyclip,ma2025aa}.
These methods typically enhance anomaly sensitivity through learnable prompts or lightweight adaptation modules. For 3D anomaly detection, CLIP is mainly employed through multi-view projection frameworks that convert point clouds into depth images and reuse the CLIP visual encoder.
Also, recent studies investigate zero-shot point cloud anomaly detection by combining multi-view projection with prompting strategies or auxiliary supervised data to improve generalization across categories \cite{cheng2024towards}.
In few-shot settings, CLIP-based 3D methods may further introduce adapter training or synthetic anomaly generation to enhance anomaly discrimination \cite{zuo2024clip3d}.

\noindent\textbf{Positioning of our work.}
Different from prompt-driven or training-based CLIP adaptations \cite{zhou2023anomalyclip,ma2025aa}, and also different from conventional reconstruction/distillation pipelines requiring training on normal training data \cite{masuda2021toward,qin2023teacher}, our method focuses on a \emph{training-free cross-category setting}: it directly uses multi-view realistic depth map projection \cite{zhu2023pointclip} and frozen CLIP visual features to compute anomaly scores from a few normal references, without any category-wise fine-tuning or additional training data preparation.
\begin{figure*}[t]
    \centering
    \includegraphics[width=0.95\textwidth]{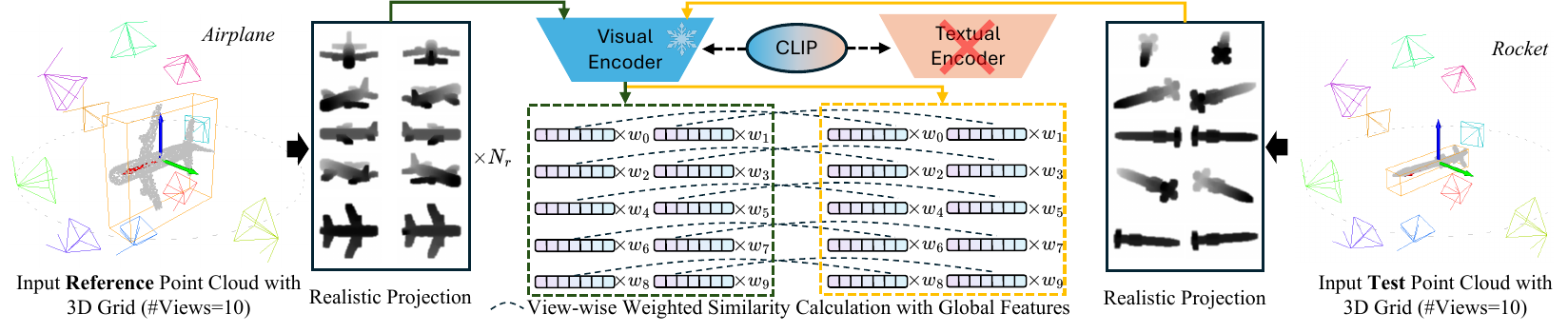}
    \caption{
    Overview of proposed DMP-3DAD. Given reference point clouds and a test point cloud, we generate realistic multi-view projections using a fixed 3D view grid.
    The projected images are encoded by the CLIP visual encoder only. View-wise weighted similarities between reference and test features are computed and aggregated to produce the final anomaly score.
    }
    \label{fig:method_overview}
\end{figure*}

\section{Method}
\subsection{Problem Definition}
\label{sec:problem}

We study the problem of \emph{cross-category 3D anomaly detection} for point clouds under a one-class and few-shot setting.
Let $\mathcal{X} \subset \mathbb{R}^{N \times 3}$ denote the space of 3D point clouds, where each sample $X \in \mathcal{X}$ consists of $N$ 3D points.
We assume that point clouds are drawn from an unknown distribution associated with an object category. Let $\mathcal{C}$ denote the set of all object categories, and let $c^{\ast} \in \mathcal{C}$ be a target (normal) category.
Point clouds belonging to $c^{\ast}$ are regarded as normal samples, while point clouds from any other category $c \in \mathcal{C} \setminus \{c^{\ast}\}$ are treated as anomalies.
This definition captures the cross-category nature of the problem, where anomalies correspond to samples from unseen categories.

During reference construction, we are given only a small set of normal point clouds from the target category:
\begin{equation}
\mathcal{R} = \{ X^r_i \mid X^r_i \sim p(X \mid c^{\ast}),\; i = 1, \dots, N_r \},
\end{equation}
where $N_r$ is small (e.g., $N_r \in \{1,3,5\}$).
No anomalous samples and category label information from $\mathcal{C}$ are available during reference construction.

At test time, the model receives an unseen point cloud $X^t \in \mathcal{X}$ drawn from an unknown category.
The goal is to determine whether $X^t$ belongs to the target category $c^{\ast}$ or should be rejected as an anomaly.
Formally, we seek an anomaly scoring function $s: \mathcal{X} \rightarrow \mathbb{R}$, such that samples from the target category yield lower anomaly scores than samples from unseen categories. For decision, given a threshold $\tau$, a test sample $X^t$ is classified as anomalous if $s(X^t) > \tau$.
The threshold $\tau$ can be determined using the normal reference set $\mathcal{R}$.
Importantly, the anomaly scoring function $s(\cdot)$ is constructed without any training or optimization procedure and does not depend on the specific choice of the target category.

\subsection{Multi-view Realistic Projection}
\label{sec:projection}

Directly applying pretrained 2D visual models to raw 3D point clouds is non-trivial due to the large domain gap between irregular 3D geometry and natural images.
A common strategy to bridge this gap is to project point clouds into multiple 2D views and reuse powerful 2D backbones.
However, naive projection methods often produce sparse and noisy depth maps that differ significantly from the visual statistics of natural images, which limits the effectiveness of pretrained models such as CLIP. To address this issue, PointCLIP V2 introduces a realistic projection scheme that generates dense and visually coherent depth maps from point clouds \cite{zhu2023pointclip}.
Instead of directly projecting points onto image planes, the method first voxelizes the point cloud into a 3D grid and then applies densification and smoothing operations.
By preserving occlusion-aware depth ordering and enforcing local spatial continuity, the resulting depth maps exhibit smoother surfaces and more realistic point distributions, significantly narrowing the domain gap between projected 3D data and natural images.

In this work, we adopt the realistic projection strategy from PointCLIP V2 as a core component of our framework.
For each input point cloud, we render a fixed number of depth maps from predefined viewpoints, forming a multi-view representation that captures complementary geometric information.
These depth maps serve as CLIP-compatible visual inputs and enable robust feature extraction without introducing any learnable projection parameters.

\subsection{CLIP-based Visual Feature Extraction}
\label{sec:clip_feature}

As shown in Fig. \ref{fig:method_overview}, given the projected multi-view realistic depth maps, we extract visual representations using a pretrained CLIP visual encoder.
Specifically, we employ the vision backbone of CLIP (e.g., ViT-B/16) and discard the text encoder entirely.
The CLIP model is kept frozen throughout the entire process, ensuring that no training or fine-tuning is required for different object categories. For each input point cloud, the multi-view projection module produces $V$ depth images from predefined viewpoints.
Each depth image is resized to match the input resolution of the CLIP visual encoder and processed independently.
Let $\phi(\cdot)$ denote the CLIP visual encoder.
For the $v$-th view of a point cloud $X$, we obtain a view-level feature vector
\begin{equation}
\mathbf{f}_v = \phi(I_v) \in \mathbb{R}^{C},
\end{equation}
where $I_v$ is the projected depth image of view $v$, and $C$ denotes the feature dimension determined by the CLIP backbone.
Each view-level feature is $\ell_2$-normalized to ensure scale consistency across views. To form a global representation for a point cloud, we concatenate the normalized view-level features from all views:
\begin{equation}
\mathbf{f}(X) = [\mathbf{f}_1, \mathbf{f}_2, \dots, \mathbf{f}_V] \in \mathbb{R}^{V \times C}.
\end{equation}

We intentionally do not employ the CLIP textual encoder in our framework.
In cross-category 3D anomaly detection, anomalies correspond to samples from unseen categories rather than semantic irregularities that can be reliably described by language.
Introducing text prompts or category names would implicitly inject category-level supervision and fundamentally alter the problem setting.
Instead, we rely solely on the pretrained CLIP visual encoder as a generic and robust feature extractor, whose representations capture object-level shape and structural consistency across viewpoints.
This design preserves the training-free nature of our approach and avoids any form of category or prompt leakage.

\begin{figure}[t]
    \centering
    \includegraphics[width=0.8\linewidth]{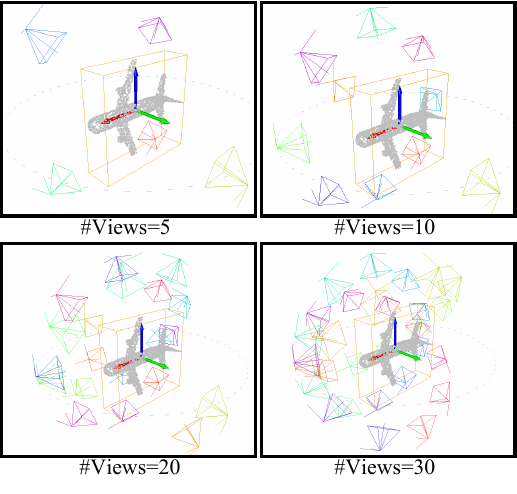}
    \caption{
    Multi-view camera configurations for point cloud projection with different numbers of views (5, 10, 20, and 30).
    Camera frustums indicate both viewpoint locations and viewing directions toward the object center.
    }
    \label{fig:multi_view_setup}
\end{figure}

\subsection{View Reliability Weighting}
Not all projected views contribute equally to anomaly detection, as some views may contain sparse or partially occluded depth information.
To account for this, we estimate a view reliability weight based on the proportion of valid depth pixels in each projected image.

Let $I_{i,v} \in [0,1]^{H \times W}$ denote the normalized depth image of the $v$-th view projected from the $i$-th reference point cloud, where pixel intensity values are scaled to $[0,1]$.
Since background regions are rendered as pure white, we define a pixel as valid if its intensity is smaller than a threshold $\gamma \in (0,1)$.
Accordingly, we construct a binary validity mask $M_{i,v} \in \{0,1\}^{H \times W}$ as
\begin{equation}
M_{i,v}(x,y) =
\begin{cases}
1, & I_{i,v}(x,y) < \gamma, \\
0, & \text{otherwise}.
\end{cases}
\end{equation}

The reliability score of view $v$ for reference sample $i$ is then computed as the ratio of valid pixels:
\begin{equation}
r_{i,v} = \frac{1}{HW} \sum_{x=1}^{H} \sum_{y=1}^{W} M_{i,v}(x,y).
\end{equation}
We obtain a global view reliability weight by averaging across the reference set:
\begin{equation}
w_v = \frac{1}{|\mathcal{R}|} \sum_{i=1}^{|\mathcal{R}|} r_{i,v},
\end{equation}
where $w_v \in [0,1]$ reflects the reliability of the $v$-th view.

These view weights are used to modulate the contribution of each view during similarity computation, reducing the influence of unreliable projections.
The entire feature extraction pipeline relies solely on pretrained CLIP visual representations and deterministic geometric projections.
No learnable parameters, auxiliary data, or category-specific adaptation are introduced.

The multi-view camera configurations are predefined and deterministic.
As illustrated in Fig.~\ref{fig:multi_view_setup}, viewpoints are uniformly distributed around the object center and oriented toward the centroid.
Different numbers of views (5, 10, 20, and 30) correspond to increasingly dense angular coverage, realized by combining horizontal ring views with additional elevated and polar viewpoints.
This design enables scalable and consistent multi-view projection across all categories without introducing any learnable parameters or data-dependent view selection.

\subsection{Anomaly Scoring via Feature Similarity}
\label{sec:anomaly_scoring}

Given the multi-view visual features extracted by the CLIP visual encoder, we perform anomaly detection by measuring feature dissimilarity between a test sample and the normal reference set. Let $\mathbf{f}_v(X) \in \mathbb{R}^{C}$ denote the $\ell_2$-normalized feature of the $v$-th view of point cloud $X$, as defined in Sec.~\ref{sec:clip_feature}.
For a test point cloud $X^t$ and a reference point cloud $X_i^r \in \mathcal{R}$, we compute a view-wise weighted distance as
\begin{equation}
d(X^t, X_i^r)
=
\sum_{v=1}^{V}
w_v \, \big\lVert \mathbf{f}_v(X^t) - \mathbf{f}_v(X_i^r) \big\rVert_2,
\end{equation}
where $w_v$ denotes the reliability weight of the $v$-th view introduced in Sec.~\ref{sec:clip_feature}.
This formulation emphasizes reliable views while suppressing noisy or sparsely projected ones.

To aggregate distances across multiple reference samples, we define the anomaly score of a test point cloud $X^t$ as
\begin{equation}
s(X^t) = \sum_{X_i^r \in \mathcal{R}} d(X^t, X_i^r).
\end{equation}
A higher anomaly score indicates larger deviation from the normal reference set and thus a higher likelihood of being anomalous.
In practice, alternative aggregation strategies such as minimum or average distance can also be applied, but we adopt summation for its robustness in few-shot settings. The resulting anomaly score $s(\cdot)$ directly corresponds to the decision rule defined in Sec.~\ref{sec:problem}.

\subsection{Discussion: Difference from Zero-shot CLIP-based Methods}
\label{sec:discussion_zs}

Recent works explore zero-shot anomaly detection by leveraging CLIP with multi-view projection for 3D data \cite{cheng2024towards,zuo2024clip3d}.
Although sharing the goal of avoiding category-specific training, these methods rely on different assumptions from the training-free cross-category setting considered in this work. However,
zero-shot CLIP-based approaches introduce textual supervision, prompt engineering, or auxiliary anomaly datasets, often combined with adapter training, to align visual features with semantic cues.
Such designs implicitly inject semantic priors and require additional data preparation or optimization.

In contrast, our method operates under a stricter formulation of one-class classification.
We assume access only to a small set of normal reference point clouds and do not employ the CLIP textual encoder, prompts, or auxiliary anomaly data.
Anomaly detection is performed purely via feature similarity in the CLIP visual embedding space.
The anomaly scoring function is fully deterministic, category-agnostic, and remains unchanged when the target category varies.
As a result, our approach provides a simple and robust alternative for practical few-shot cross-category 3D anomaly detection, complementing existing zero-shot CLIP-based methods.

\section{Experiment}
\subsection{Experimental Setup}
\label{sec:exp_setup}

\noindent\textbf{Datasets.}
Following the previous works \cite{masuda2021toward, qin2023teacher}, we evaluate the proposed method on the ShapeNetPart dataset \cite{yi2017large}.
ShapeNetPart is a subset of ShapeNet that provides object-level point clouds across multiple semantic categories.
Following the official dataset protocol, we adopt the predefined training and test splits, which contain 12{,}137 training samples and 2{,}874 test samples, respectively. Each object is represented as a point cloud consisting of 2{,}048 points, which are randomly sampled from the underlying 3D shape surface.
All point clouds are normalized and consistently oriented, enabling fair comparison across categories.
The dataset includes 16 object categories, such as airplane, chair, table, and lamp, covering diverse global geometries and structural variations.

\noindent\textbf{Few-shot reference setting.}
To simulate realistic few-shot scenarios, we randomly sample a small number of normal point clouds from the target category as reference samples.
Unless otherwise specified, we consider $N_r \in \{1, 3, 5\}$ normal references.
No anomalous samples are used during reference construction, and labels are all unavailable for scoring.

\noindent\textbf{Multi-view projection.}
Each point cloud is projected into a fixed number of realistic depth maps using the multi-view projection strategy described in Sec.~\ref{sec:projection}.
We evaluate different numbers of views ($V=5, 10, 20, 30$) to study the effect of view number.
Unless otherwise stated, we use $V=10$ views as a trade-off between computational efficiency and performance.

\noindent\textbf{Evaluation protocol and metric.}
We use the area under the receiver operating characteristic curve (AUROC) as the primary evaluation metric following \cite{qin2023teacher,masuda2021toward}, where higher scores indicate a higher likelihood of being anomalous.
For robustness, all experiments are repeated with 10 different random seeds for reference sample selection, and the average AUROC is reported.
AUROC is computed for each target category independently, and the mean AUROC across all categories is used as the final performance measure.

\begin{table*}[t]
    \centering
    \caption{Category-wise AUC-ROC (\%) comparison for few-shot anomaly detection.
The target category (normal class) is used for reference, and all other categories are regarded as anomalies.
Results are reported with 1, 3, and 5 reference samples.
\textbf{The reconstruction-based and knowledge-distillation-based methods require category-wise reference, whereas our method does not require training.}
Mean and standard deviation are computed over 10 random runs.
$\uparrow$ ($\downarrow$) indicates that higher (lower) values are better.
Best results are shown in bold.}
    \label{tab:tab2}
    
    \setlength{\tabcolsep}{3pt}
    \renewcommand{\arraystretch}{1.2}
    \scriptsize   
    
    \begin{tabular}{
l
>{\columncolor{sampleOne}}c
>{\columncolor{sampleThree}}c
>{\columncolor{sampleFive}}c
>{\columncolor{sampleOne}}c
>{\columncolor{sampleThree}}c
>{\columncolor{sampleFive}}c
>{\columncolor{sampleOne}}c
>{\columncolor{sampleThree}}c
>{\columncolor{sampleFive}}c
}
        \toprule
        & \multicolumn{3}{c}{Reconstruction-based~\cite{masuda2021toward}}
        & \multicolumn{3}{c}{Knowledge-distillation-based~\cite{qin2023teacher}}
        & \multicolumn{3}{c}{DMP-3DAD (ours)}\\
        \cmidrule(lr){2-4}
        \cmidrule(lr){5-7}
        \cmidrule(lr){8-10}
        \multicolumn{1}{c}{Category (\#tests)}
        & 1 sample & 3 samples & 5 samples
        & 1 sample & 3 samples & 5 samples
        & 1 sample & 3 samples & 5 samples\\
        \midrule
        Airplane (341)
        & 87.60 $\pm$ 5.99  & 96.07 $\pm$ 1.07  & 97.39 $\pm$ 0.48
        & 97.41 $\pm$ \textbf{1.01}  & {98.59 $\pm$ 0.18} & 98.29 $\pm$ 0.53
        & \textbf{98.73}$\pm$2.22 & \textbf{99.71}$\pm$\textbf{0.11} & \textbf{99.75}$\pm$\textbf{0.08} \\
        Bag (14)
        & 47.09 $\pm$ 8.73  & 52.41 $\pm$ 5.86  & 58.70 $\pm$ 5.79
        & \textbf{96.88} $\pm$ \textbf{3.73}  & \textbf{98.23} $\pm$ \textbf{2.13}  & \textbf{99.94} $\pm$ \textbf{0.08}
        & 90.26$\pm$6.57 & 94.18$\pm$3.98 & 94.85$\pm$3.51 \\
        Cap (11)
        & 38.71 $\pm$ 5.70  & 45.22 $\pm$ 3.82  & 46.31 $\pm$ 6.38
        & 90.96 $\pm$ 5.54  & {94.96 $\pm$ 3.05} & 94.13 $\pm$ 2.94
        & \textbf{97.85}$\pm$\textbf{1.30} & \textbf{98.35}$\pm$\textbf{1.35} & \textbf{98.18}$\pm$\textbf{2.20} \\
        Car (158)
        & 62.28 $\pm$ 3.27  & 64.12 $\pm$ 2.06  & 65.14 $\pm$ 1.81
        & \textbf{99.33} $\pm$ \textbf{0.27}  & {99.36 $\pm$ 0.26} & 99.31 $\pm$ 0.29
        & 99.32$\pm$0.69 & \textbf{99.69}$\pm$\textbf{0.14} & \textbf{99.75}$\pm$\textbf{0.09} \\
        Chair (704)
        & 49.20 $\pm$ 4.27  & 53.38 $\pm$ 2.96  & 55.38 $\pm$ 1.29
        & \textbf{95.16} $\pm$ \textbf{2.20}  & \textbf{98.54} $\pm$ \textbf{0.64}  & \textbf{98.72} $\pm$ \textbf{0.18}
        & 84.34$\pm$7.56 & 93.10$\pm$3.52 & 93.33$\pm$2.30 \\
        Earphone (14)
        & 38.78 $\pm$ \textbf{7.61}  & 45.36 $\pm$ 7.26  & 43.64 $\pm$ 3.21
        & 81.97 $\pm$ 23.45 & {91.31 $\pm$ 3.07} & 90.19 $\pm$ 2.22
        & \textbf{89.08}$\pm$21.45 & \textbf{99.95}$\pm$\textbf{0.16} & \textbf{100.00}$\pm$\textbf{0.00} \\
        Guitar (159)
        & 71.75 $\pm$ 3.45  & 76.13 $\pm$ 3.74  & 77.59 $\pm$ 2.57
        & {98.65 $\pm$ \textbf{0.54}} & 97.66 $\pm$ 1.13 & 98.39 $\pm$ 0.69
        & \textbf{97.53}$\pm$1.93 & \textbf{98.99}$\pm$\textbf{0.86} & \textbf{99.22}$\pm$\textbf{0.30} \\
        Knife (80)
        & 66.46 $\pm$ 4.20  & 70.49 $\pm$ 2.31  & 71.79 $\pm$ 0.91
        & \textbf{95.18} $\pm$ \textbf{1.75}  & \textbf{95.33} $\pm$ \textbf{2.22}  & {\textbf{96.72} $\pm$ \textbf{1.08}}
        & 88.09$\pm$15.13 & 91.84$\pm$4.18 & 93.74$\pm$2.69 \\
        Lamp (286)
        & 53.09 $\pm$ \textbf{5.02}  & 58.68 $\pm$ \textbf{2.55}  & {62.20 $\pm$ 3.68}
        & 56.13 $\pm$ 6.16  & 60.76 $\pm$ 8.34  & 61.22 $\pm$ 5.84
        & \textbf{68.48}$\pm$11.70 & \textbf{82.65}$\pm$5.95 & \textbf{83.20}$\pm$\textbf{2.90} \\
        Laptop (83)
        & 67.36 $\pm$ 4.52  & 69.08 $\pm$ 2.24  & 70.24 $\pm$ 2.58
        & {\textbf{98.89} $\pm$ 0.16} & 98.78 $\pm$ 0.32 & 98.69 $\pm$ 0.29
        & 98.88$\pm$\textbf{0.05} & \textbf{98.89}$\pm$\textbf{0.06} & \textbf{98.90}$\pm$\textbf{0.05} \\
        Motorbike (51)
        & 82.55 $\pm$ \textbf{2.75}  & 87.66 $\pm$ 1.28  & 88.09 $\pm$ 2.56
        & 90.63 $\pm$ 26.57 & 99.27 $\pm$ 0.72 & {99.35 $\pm$ 0.60}
        & \textbf{98.19}$\pm$3.11 & \textbf{99.62}$\pm$\textbf{0.66} & \textbf{99.68}$\pm$\textbf{0.52} \\
        Mug (38)
        & 44.04 $\pm$ 5.67  & 46.94 $\pm$ 2.44  & 48.55 $\pm$ 3.82
        & 99.60 $\pm$ \textbf{0.49}  & {99.73 $\pm$ 0.37} & 99.35 $\pm$ 0.57
        & \textbf{99.68}$\pm$0.50 & \textbf{99.96}$\pm$\textbf{0.06} & \textbf{99.96}$\pm$\textbf{0.06} \\
        Pistol (44)
        & 64.85 $\pm$ 4.70  & 71.77 $\pm$ 4.41  & 78.57 $\pm$ 4.44
        & {\textbf{99.08} $\pm$ \textbf{0.66}} & 98.86 $\pm$ \textbf{0.51} & 98.80 $\pm$ \textbf{0.48}
        & 98.67$\pm$1.43 & \textbf{99.41}$\pm$1.14 & \textbf{99.53}$\pm$0.65 \\
        Rocket (12)
        & 55.25 $\pm$ 2.97  & 55.95 $\pm$ 5.69  & 59.67 $\pm$ 5.05
        & {\textbf{96.63} $\pm$ \textbf{1.44}} & \textbf{95.73} $\pm$ \textbf{2.01} & \textbf{96.23} $\pm$ \textbf{2.93}
        & 89.38$\pm$6.94 & 93.68$\pm$4.98 & 94.24$\pm$4.10 \\
        Skateboard (31)
        & 47.84 $\pm$ 5.19  & 57.56 $\pm$ 5.08  & 57.29 $\pm$ 6.62
        & {\textbf{96.16} $\pm$ \textbf{1.20}} & 95.64 $\pm$ \textbf{1.06} & 96.07 $\pm$ \textbf{1.11}
        & 94.59$\pm$3.66 & \textbf{97.17}$\pm$2.42 & \textbf{98.11}$\pm$1.37 \\
        Table (848)
        & 47.39 $\pm$ 14.48 & 61.97 $\pm$ 14.66 & 78.78 $\pm$ 4.56
        & 83.46 $\pm$ 8.17  & 89.77 $\pm$ 2.34  & {90.40 $\pm$ 1.05}
        & \textbf{86.82}$\pm$\textbf{5.98} & \textbf{90.03}$\pm$\textbf{1.90} & \textbf{90.65}$\pm$\textbf{2.00} \\
        \midrule
        Avg. AUC-ROC $\uparrow$
        & 57.77 & 63.30 & 66.21
        & 92.26 & 94.53 & {94.74}
        & \textbf{92.49}   & \textbf{96.08}  & \textbf{96.44} \\
        Avg. Std. Dev. $\downarrow$
        & 5.53  & 4.21  & 3.48
        & \textbf{5.21}  & \textbf{1.77}  & \textbf{1.30}
        & 5.64    & 1.97    & 1.43 \\
        \bottomrule
    \end{tabular}
\end{table*}

\subsection{Comparison with State-of-the-Art Methods}
\label{sec:comparison}

Tab.~\ref{tab:tab2} reports the category-wise AUC-ROC comparison between the proposed method and representative state-of-the-art approaches for few-shot cross-category 3D anomaly detection.
We compare against a reconstruction-based method \cite{masuda2021toward} and a knowledge-distillation-based method \cite{qin2023teacher}, both of which require category-wise training using normal reference samples.
In contrast, our approach is entirely training-free and does not involve any optimization or category-specific adaptation. Overall, the proposed method achieves the best average performance across all few-shot settings. In particular, our method attains mean AUC-ROC scores of 92.49\%, 96.08\%, and 96.44\% with 1, 3, and 5 reference samples, respectively, outperforming the compared methods in all three settings.
The performance becomes better as the number of reference samples increases, indicating that our similarity-based formulation can effectively exploit additional normal exemples without retraining. At the category level, our method consistently achieves strong results on most object classes, especially those with complex global geometries such as \emph{airplane}, \emph{earphone} and \emph{guitar}.
While knowledge-distillation-based methods can achieve competitive performance on certain categories, they rely on category-specific training to explicitly adapt the feature space to the normal data distribution.
Such adaptation can contribute to reduced variance across different reference selections.
In contrast, our method operates in a fully training-free manner and does not modify the underlying feature space according to the available normal samples.
As a result, the anomaly scores are directly determined by the similarity relationships in the frozen CLIP embedding space, which naturally leads to relatively higher variance across different reference sets.
Nevertheless, this behavior is expected and acceptable, as it reflects the absence of optimization or category-dependent fitting, while still maintaining strong average performance across categories.

\subsection{Ablation Studies}
\label{sec:ablation}

\noindent\textbf{Effect of View Weighting.}
We first analyze the impact of the view reliability weighting threshold $\gamma$, which determines whether a pixel is considered valid based on its depth intensity.
Tab.~\ref{tab:weighting_ablation} shows that the proposed method is insensitive to the exact choice of $\gamma$.
Across a wide range of thresholds, the average AUC-ROC remains stable for all few-shot settings.
Slightly better performance and lower variance are consistently observed when $\gamma$ is set between 0.2 and 0.3.
Based on this observation, we set $\gamma=0.2$ in all experiments.
These results indicate that view weighting provides a robust mechanism for suppressing unreliable projections without requiring careful parameter tuning.

\noindent\textbf{Effect of Distance Metrics.}
We evaluate different distance metrics for feature similarity computation, including cosine distance, Euclidean distance, and Manhattan distance.
As reported in Tab.~\ref{tab:distance_ablation}, all three metrics yield comparable performance, confirming that the learned CLIP visual features are well-structured in the embedding space.
Among them, Euclidean distance achieves the best overall performance, especially in the 3- and 5-shot settings, and is therefore adopted as the default choice in our framework.

\noindent\textbf{Effect of CLIP Backbones.}
Tab.~\ref{tab:backbone_ablation} compares different CLIP visual backbones under the same experimental protocol.
Vision Transformer backbones consistently outperform convolutional architectures, indicating their stronger capacity for capturing global geometric patterns from multi-view depth images.
In particular, ViT-B/32 achieves the best trade-off between accuracy and stability across all few-shot settings.
Unless otherwise stated, ViT-B/32 is used in all experiments.

\noindent\textbf{Effect of the Number of Views.}
We further study the influence of the number of projected views in Tab.~\ref{tab:view_ablation}.
Using only 5 views leads to a noticeable performance drop due to insufficient coverage of object geometry.
Increasing the number of views from 10 to 20 yields marginal improvements, while further increasing to 30 views does not bring additional gains and may slightly degrade performance due to redundant or less informative projections.
Considering both accuracy and computational efficiency, we use 10 views as the default setting.

\begin{figure}[tb] 
\centering 
\includegraphics[width=1.0\linewidth]{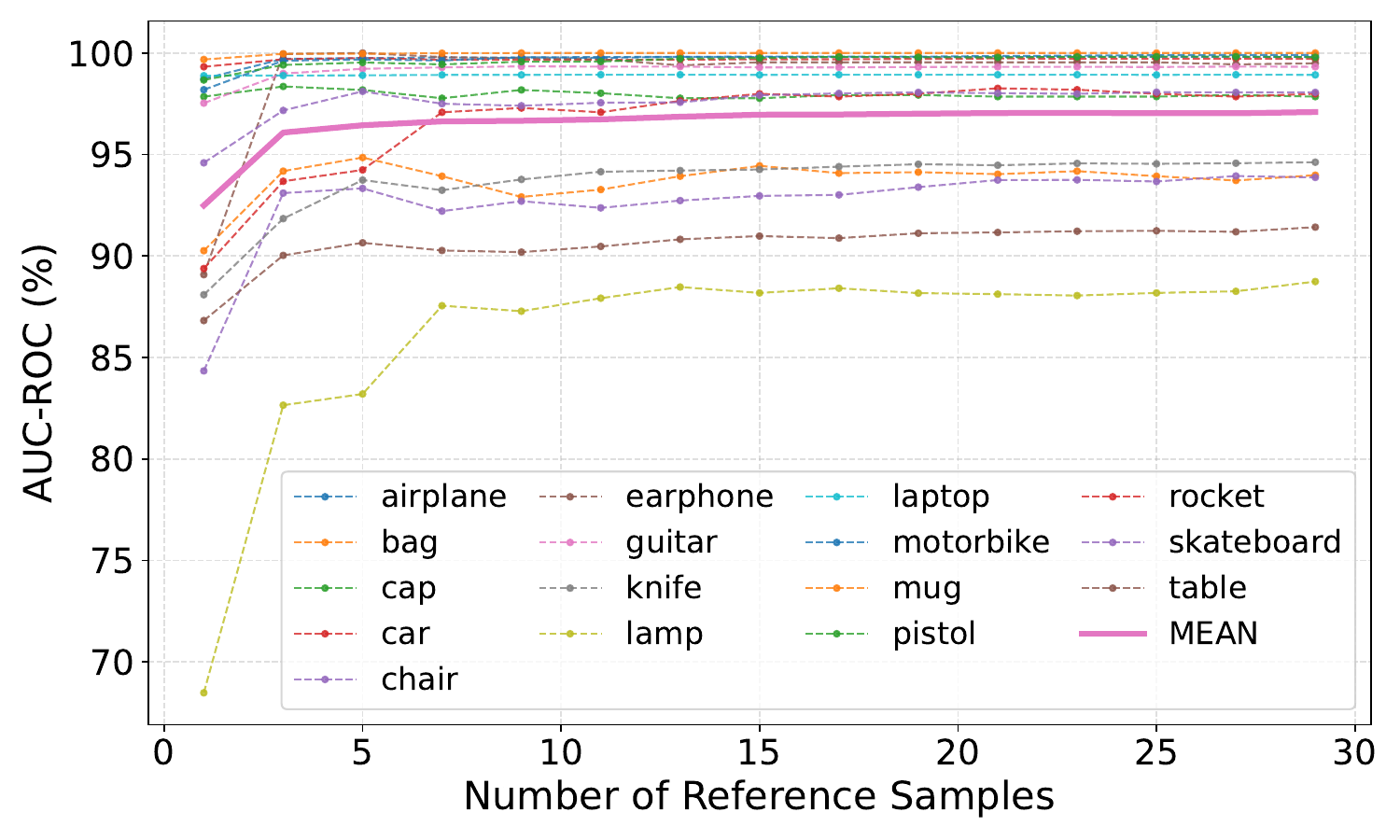} 
\caption{Category-wise AUC-ROC (\%) with varying numbers of reference samples. Thin dashed curves correspond to individual object categories, while the bold solid curve shows the mean performance averaged over all categories.}
\label{fig:ref_auc} 
\end{figure}

\begin{figure}[tb] 
\centering 
\includegraphics[width=1.0\linewidth]{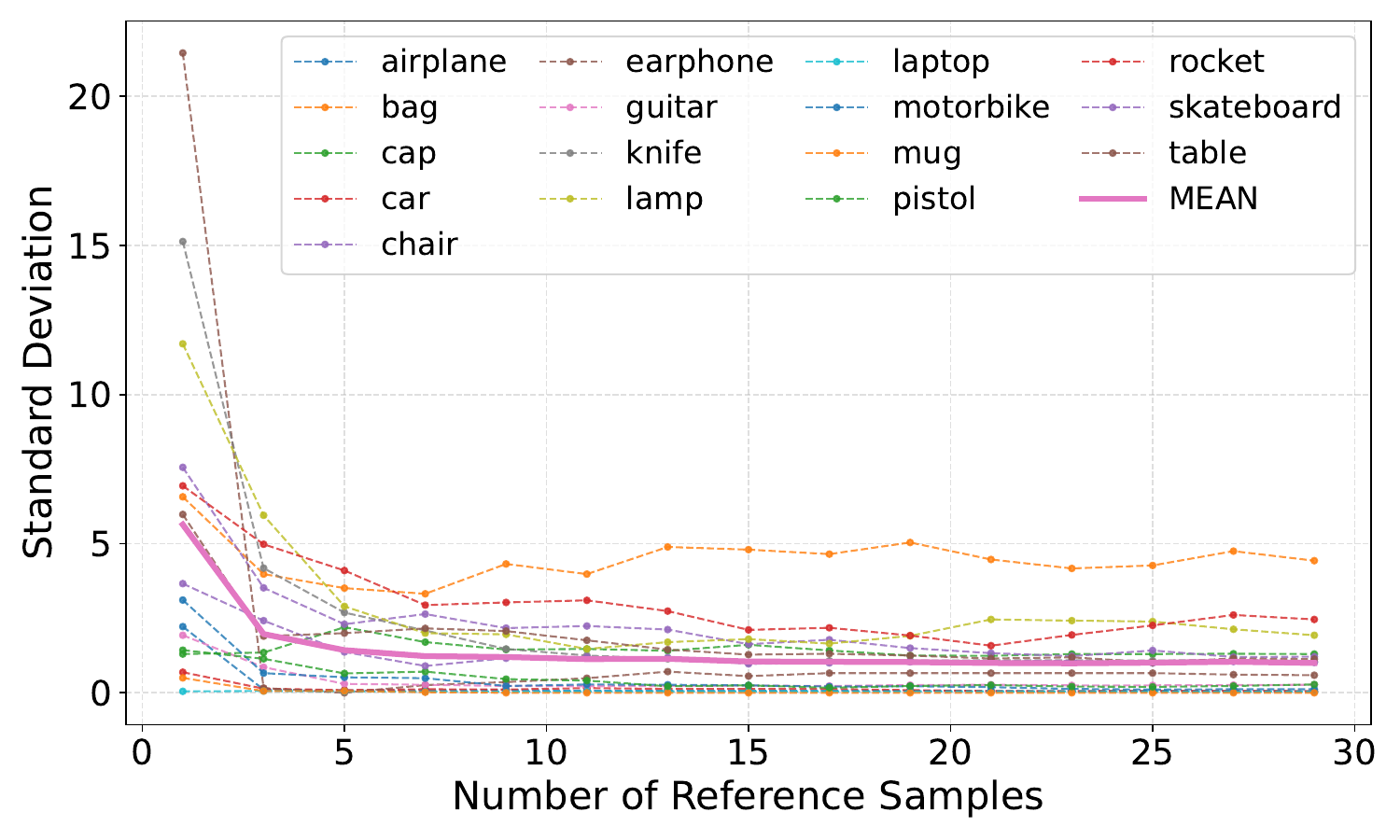} 
\caption{Category-wise standard deviation with varying numbers of reference samples. Thin dashed curves correspond to individual object categories, while the bold solid curve shows the mean performance averaged over all categories.}
\label{fig:ref_std} 
\end{figure}

\noindent\textbf{Effect of the Number of References.}
We analyze the influence of the number of normal reference samples on anomaly detection performance.
Fig.~\ref{fig:ref_auc} reports the category-wise AUC-ROC as the reference size increases, while Fig.~\ref{fig:ref_std} shows the corresponding standard deviation over 10 random runs.
With only a single reference sample, the proposed method already achieves strong performance across most categories, demonstrating its effectiveness in extreme few-shot scenarios.
As more reference samples are introduced, the AUC-ROC consistently improves and gradually saturates, indicating that the CLIP-based visual embeddings provide a stable representation for cross-category similarity comparison.
Meanwhile, the standard deviation decreases significantly with increasing reference size, especially for categories with larger intra-class variability.
These results suggest that additional normal references mainly contribute to stabilizing the similarity estimation rather than fundamentally altering the decision boundary.
Overall, the proposed training-free framework exhibits robust and predictable behavior with respect to the number of reference samples.

\noindent\textbf{Component Analysis.}
Finally, Tab.~\ref{tab:component_ablation} presents an ablation study on the main components of the proposed framework.
Removing multi-view projection (MP) leads to a substantial performance degradation, highlighting its necessity.
Introducing view-wise similarity (VS) significantly improves performance, and incorporating view weighting (VW) further enhances robustness and reduces variance.
The full model consistently achieves the best overall performance, demonstrating the complementary roles of all components.

\section{Analysis of Failure Cases}
\label{sec:failure}

\begin{figure*}[t]
    \centering
    \includegraphics[width=\linewidth]{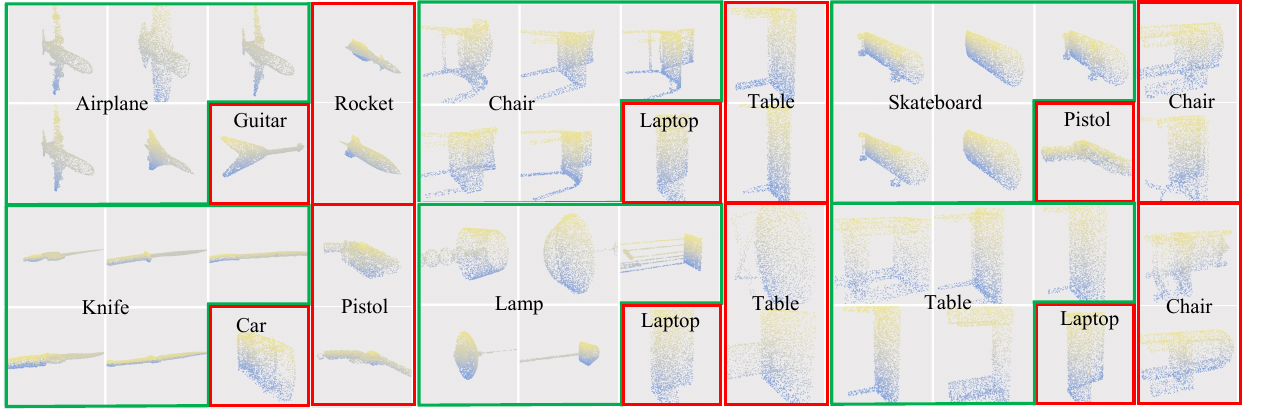}
    \caption{
    Failure cases of the proposed anomaly detection method.
    Green boxes indicate reference samples from the normal class,
    while red boxes denote anomalous samples from different classes
    that are incorrectly classified as normal.
    }
    \label{fig:failure_cases}
\end{figure*}

Fig.~\ref{fig:failure_cases} shows representative failure cases of the proposed method, where anomalous samples from unseen categories are incorrectly classified as normal.
A typical failure occurs when anomalous objects share strong global geometric similarity with the target category.
For example, objects such as \emph{rocket} or \emph{guitar} may resemble \emph{airplane} under certain viewpoints due to their elongated shapes, while \emph{laptop} samples can be confused with \emph{chair} or \emph{table} because of similar planar structures.
In such cases, multi-view depth projections yield visually similar silhouettes, resulting in small feature distances in the CLIP visual embedding space.

Another source of error stems from view-dependent ambiguity.
Although multi-view projection reduces viewpoint sensitivity, discriminative cues may only appear in a limited number of views.
When most views exhibit high similarity to normal references, the aggregated similarity score may still favor normality, especially for structurally simple or symmetric categories such as \emph{table} and \emph{chair}.

These failure cases reflect an inherent limitation of training-free, similarity-based detection, which does not explicitly model fine-grained local geometry or category-specific semantics.
Nevertheless, our method enables strong generalization without category-wise training.
Incorporating additional local geometric cues or adaptive view strategies may further improve discrimination in such challenging cases, which we leave for future work.

\begin{table}[t]
    \centering
    \caption{Effect of weighting threshold on anomaly detection performance.
    Results are reported in terms of average AUC-ROC (\%) and average standard deviation
    over 10 random runs with 1, 3, and 5 reference samples.}
    \label{tab:weighting_ablation}

    \setlength{\tabcolsep}{6pt}
    \renewcommand{\arraystretch}{1.0}
    \small

    \begin{tabular}{lcccccc}
        \toprule
        & \multicolumn{6}{c}{Number of reference samples} \\
        \cmidrule(lr){2-7}
        Weighting
        & \multicolumn{2}{c}{1 sample}
        & \multicolumn{2}{c}{3 samples}
        & \multicolumn{2}{c}{5 samples} \\
        \cmidrule(lr){2-3}
        \cmidrule(lr){4-5}
        \cmidrule(lr){6-7}
        & Mean & Std
        & Mean & Std
        & Mean & Std \\
        \midrule
        $\gamma<0.1$
        & 92.48 & 5.48
        & 96.01 & 1.98
        & \textbf{96.47} & 1.43 \\
        $\gamma<0.2$
        & \textbf{92.63} & \textbf{5.34}
        & \textbf{96.07} & 1.89
        & 96.42 & 1.39 \\
        $\gamma<0.3$
        & 92.59 & 5.46
        & 96.05 & \textbf{1.83}
        & 96.39 & \textbf{1.34} \\
        $\gamma<0.4$
        & 92.61 & 5.41
        & 95.99 & 1.88
        & 96.33 & 1.36 \\
        $\gamma<0.5$
        & 92.61 & 5.40
        & 95.92 & 1.92
        & 96.28 & 1.38 \\
        $\gamma<0.6$
        & 92.52 & 5.57
        & 95.89 & 1.92
        & 96.26 & 1.37 \\
        $\gamma<0.7$
        & 92.47 & 5.64
        & 95.90 & 1.89
        & 96.24 & 1.35 \\
        $\gamma<0.8$
        & 92.48 & 5.63
        & 95.87 & 1.91
        & 96.25 & 1.36 \\
        $\gamma<0.9$
        & 92.49 & 5.61
        & 95.86 & 1.92
        & 96.24 & 1.36 \\
        $\gamma<1.0$
        & 92.49 & 5.61
        & 95.84 & 1.86
        & 96.32 & 1.54 \\
        \bottomrule
    \end{tabular}
\end{table}

\begin{table}[t]
    \centering
    \caption{Effect of distance metrics on anomaly detection performance.
    Results are reported in terms of average AUC-ROC (\%) and average standard deviation
    over 10 random runs with 1, 3, and 5 reference samples.}
    \label{tab:distance_ablation}

    \setlength{\tabcolsep}{6pt}
    \renewcommand{\arraystretch}{1.0}
    \small

    \begin{tabular}{lcccccc}
        \toprule
        & \multicolumn{6}{c}{Number of reference samples} \\
        \cmidrule(lr){2-7}
        Metric
        & \multicolumn{2}{c}{1 sample}
        & \multicolumn{2}{c}{3 samples}
        & \multicolumn{2}{c}{5 samples} \\
        \cmidrule(lr){2-3}
        \cmidrule(lr){4-5}
        \cmidrule(lr){6-7}
        & Mean & Std
        & Mean & Std
        & Mean & Std \\
        \midrule
        Cosine
        & 92.61 & 5.32
        & 95.75 & 1.93
        & 96.10 & \textbf{1.35} \\
        Euclidean
        & 92.63 & 5.34
        & \textbf{96.07} & \textbf{1.89}
        & \textbf{96.42} & 1.39 \\
        Manhattan
        & \textbf{92.64} & \textbf{5.25}
        & 96.05 & 1.95
        & \textbf{96.42} & 1.40 \\
        \bottomrule
    \end{tabular}
\end{table}

\begin{table}[t]
    \centering
    \caption{Effect of backbone architectures on anomaly detection performance.
    Results are reported in terms of average AUC-ROC (\%) and standard deviation
    over 10 random runs with 1, 3, and 5 reference samples.}
    \label{tab:backbone_ablation}

    \setlength{\tabcolsep}{6pt}
    \renewcommand{\arraystretch}{1.1}
    \small

    \begin{tabular}{lcccccc}
        \toprule
        & \multicolumn{6}{c}{Number of reference samples} \\
        \cmidrule(lr){2-7}
        Backbone
        & \multicolumn{2}{c}{1 sample}
        & \multicolumn{2}{c}{3 samples}
        & \multicolumn{2}{c}{5 samples} \\
        \cmidrule(lr){2-3}
        \cmidrule(lr){4-5}
        \cmidrule(lr){6-7}
        & Mean & Std
        & Mean & Std
        & Mean & Std \\
        \midrule
        ViT-B/16
        & 92.10 & 5.59
        & 95.64 & 1.96
        & 96.06 & 1.61 \\
        ViT-B/32
        & \textbf{92.63} & \textbf{5.34}
        & \textbf{96.07} & \textbf{1.89}
        & \textbf{96.42} & \textbf{1.39} \\
        RN50
        & 88.34 & 7.47
        & 93.03 & 3.42
        & 93.70 & 2.63 \\
        RN101
        & 89.68 & 7.05
        & 94.81 & 2.94
        & 95.31 & 2.20 \\
        RN50$\times$4
        & 90.30 & 7.15
        & 94.56 & 3.15
        & 95.14 & 2.38 \\
        RN50$\times$16
        & 89.46 & 8.79
        & 93.55 & 3.82
        & 93.89 & 2.94 \\
        \bottomrule
    \end{tabular}
\end{table}

\begin{table}[t]
    \centering
    \caption{Effect of the number of views on anomaly detection performance.
    Results are reported in terms of average AUC-ROC (\%) and standard deviation
    over 10 random runs with 1, 3, and 5 reference samples.}
    \label{tab:view_ablation}

    \setlength{\tabcolsep}{6pt}
    \renewcommand{\arraystretch}{1.1}
    \small

    \begin{tabular}{ccccccc}
        \toprule
        & \multicolumn{6}{c}{Number of reference samples} \\
        \cmidrule(lr){2-7}
        \#Views
        & \multicolumn{2}{c}{1 sample}
        & \multicolumn{2}{c}{3 samples}
        & \multicolumn{2}{c}{5 samples} \\
        \cmidrule(lr){2-3}
        \cmidrule(lr){4-5}
        \cmidrule(lr){6-7}
        & Mean & Std
        & Mean & Std
        & Mean & Std \\
        \midrule
        5
        & 90.34 & 6.54
        & 94.50 & 2.23
        & 94.95 & 1.64 \\
        10
        & \textbf{92.63} & \textbf{5.34}
        & 96.07 & \textbf{1.89}
        & 96.42 & \textbf{1.39} \\
        20
        & 92.49 & 5.64
        & \textbf{96.08} & 1.97
        & \textbf{96.44} & 1.43 \\
        30
        & 91.78 & 5.88
        & 95.33 & 2.21
        & 95.60 & 1.52 \\
        \bottomrule
    \end{tabular}
\end{table}

\begin{table}[t]
    \centering
\caption{Ablation study of different components in the proposed framework. MP, VS, and VW denote Multiview Projection, View-wise Similarity, and View Weighing, respectively. Results are reported in terms of average AUC-ROC (\%) and standard deviation over 10 random runs with 1, 3, and 5 reference samples.}
    \label{tab:component_ablation}

    \setlength{\tabcolsep}{4pt}
    \renewcommand{\arraystretch}{1.1}
    \small

    \begin{tabular}{ccccccccc}
        \toprule
        \multicolumn{3}{c}{Components}
        & \multicolumn{6}{c}{Number of reference samples} \\
        \cmidrule(lr){1-3}
        \cmidrule(lr){4-9}
        MP
        & VS
        & VW
        & \multicolumn{2}{c}{1 sample}
        & \multicolumn{2}{c}{3 samples}
        & \multicolumn{2}{c}{5 samples} \\
        \cmidrule(lr){4-5}
        \cmidrule(lr){6-7}
        \cmidrule(lr){8-9}
        & & 
        & Mean & Std
        & Mean & Std
        & Mean & Std \\
        \midrule
        $\times$  & $\times$  & $\times$
        & 86.84 & 7.34
        & 91.26 & 3.01
        & 91.67 & 2.49 \\
        $\checkmark$ & $\times$  & $\times$
        & 90.75 & 5.77
        & 94.31 & 2.51
        & 94.74 & 1.68 \\
        $\checkmark$ & $\checkmark$ & $\times$
        & \textbf{92.64} & 5.35
        & 95.87 & 1.89
        & 96.23 & \textbf{1.32} \\
        $\checkmark$ & $\checkmark$ & $\checkmark$
        & 92.63 & \textbf{5.34}
        & \textbf{96.07} & \textbf{1.89}
        & \textbf{96.42} & 1.39 \\
        \bottomrule
    \end{tabular}
\end{table}

\section{Conclusion}

This paper presented DMP-3DAD, a training-free framework for cross-category 3D anomaly detection under few-shot settings.
By combining realistic multi-view depth map projection with a frozen CLIP visual encoder, anomaly detection is formulated as a feature similarity problem without requiring category-specific training or optimization.
The proposed view reliability weighting improves robustness while keeping the entire pipeline fully deterministic.
Experiments on the ShapeNetPart benchmark demonstrate that DMP-3DAD achieves state-of-the-art performance with stable behavior across different numbers of reference samples.
These results indicate that leveraging realistic projection and large pretrained visual models is an effective and practical direction for training-free 3D anomaly detection, and future work may further enhance discrimination by incorporating local geometric cues or adaptive view strategies.

\bibliographystyle{ieicetr}
\bibliography{egbib}

\end{document}